\documentclass[11pt,a4paper]{article}

\usepackage[margin=1in]{geometry}
\usepackage{amsmath,amssymb,amsthm}
\usepackage{booktabs}
\usepackage{longtable}
\usepackage{array}
\usepackage{graphicx}
\usepackage[dvipsnames,table]{xcolor}
\usepackage[colorlinks=true,linkcolor=NavyBlue,citecolor=NavyBlue,urlcolor=NavyBlue]{hyperref}
\usepackage{microtype}
\usepackage{caption}
\usepackage{subcaption}
\usepackage{listings}
\usepackage{cite}
\usepackage{parskip}
\usepackage{enumitem}
\usepackage{float}
\usepackage{setspace}
\usepackage{titlesec}
\usepackage{fancyhdr}
\usepackage{tcolorbox}
\usepackage{mdframed}
\usepackage{tikz}
\usetikzlibrary{shapes.geometric, shapes.misc, arrows.meta, positioning,
                fit, backgrounds, calc, decorations.pathreplacing}
\usepackage{environ}

\titleformat{\section}{\large\bfseries}{}{0em}{}[\titlerule]
\titleformat{\subsection}{\normalsize\bfseries}{}{0em}{}
\titleformat{\subsubsection}{\normalsize\itshape\bfseries}{}{0em}{}

\title{%
  \textbf{CIPHER: Conformer-based Inference of Phonemes}\\
  \textbf{from High-density EEG Representations}
}
\author{%
  Varshith Madishetty\\
  \small\texttt{madishettyvarshith@gmail.com}
}
\date{}

\begin{document}

\maketitle
\thispagestyle{empty}

\begin{mdframed}[backgroundcolor=gray!8, linecolor=gray!40, linewidth=0.8pt,
                 innerleftmargin=10pt, innerrightmargin=10pt,
                 innertopmargin=8pt, innerbottommargin=8pt]e
\textbf{Abstract.}
Decoding speech information from scalp EEG remains difficult due to low SNR
and spatial blurring. We present \textbf{CIPHER} (Conformer-based Inference
of Phonemes from High-density EEG Representations), a dual-pathway model
using (i) ERP features and (ii) broadband DDA coefficients. On OpenNeuro
ds006104 (24 participants, two studies with concurrent TMS), binary
articulatory tasks reach near-ceiling performance but are highly
confound-vulnerable (acoustic onset separability and TMS-target blocking).
On the primary 11-class CVC phoneme task under full Study~2 LOSO (16
held-out subjects), performance is substantially lower (real-word WER:
ERP $0.671 \pm 0.080$, DDA $0.688 \pm 0.096$), indicating limited
fine-grained discriminability. We therefore position this work as a
benchmark and feature-comparison study rather than an EEG-to-text system,
and we constrain neural-representation claims to confound-controlled evidence.
\end{mdframed}

\tableofcontents
\newpage

\thispagestyle{empty}
\vspace*{\fill}
\begin{center}
  \begin{minipage}{0.78\linewidth}
    \begin{mdframed}[backgroundcolor=gray!4, linecolor=gray!35, linewidth=0.8pt,
                 innerleftmargin=18pt, innerrightmargin=18pt,
                 innertopmargin=16pt, innerbottommargin=16pt,
                 roundcorner=4pt]
      \begin{center}
        \textit{\large Dedication}\\[10pt]
      \end{center}
      \small\setlength{\parskip}{7pt}
      This work is dedicated to the memory of my grandfather,
      \textbf{Late Tarala Narsingha Rao}, who lived with a lifelong neural
      disorder that severely limited his ability to communicate. From early
      childhood, he faced profound difficulty expressing his thoughts, needs,
      and pain. Over time, his ability to communicate diminished further, until
      he was no longer able to convey even discomfort while enduring significant
      internal suffering.

      Witnessing this silence shaped my understanding of the human cost of lost
      communication. This work is guided by the belief that restoring even a
      fragile channel of expression can transform silence into dignity, isolation
      into connection, and agency for individuals who would otherwise remain
      unheard.
    \end{mdframed}
  \end{minipage}
\end{center}
\vspace*{\fill}
\newpage

\section{Introduction}

The ability to decode speech from neural activity has profound implications
for assistive communication technologies, particularly for individuals with
severe motor impairments such as locked-in syndrome or amyotrophic lateral
sclerosis (ALS). While intracortical recordings have achieved remarkable
decoding performance---including real-time sentence-level speech
synthesis~\cite{doi:10.1056/NEJMoa2314132,willett2023,metzger2023}---the clinical invasiveness of
implanted electrode arrays severely limits their scalability. Scalp EEG
offers a non-invasive alternative with millisecond temporal resolution, but
suffers from volume conduction, low signal-to-noise ratio, and spatial
blurring, making fine-grained speech decoding substantially more challenging.

Prior work on EEG-based speech decoding has largely relied on either
traditional event-related potential (ERP) analysis or spectral feature
extraction~\cite{defossez2023,anumanchipalli2019}. ERPs capture the
phase-locked cortical response to stimulus events (e.g., the auditory
N1/P2 complex, the speech-selective N200), but by averaging across trials,
they discard the trial-by-trial temporal dynamics that carry discriminative
information. Conversely, approaches based on spectral or time-frequency
features can capture sustained oscillatory activity but may miss the
transient, nonlinear dynamics characteristic of speech processing.

In this work, we propose a dual-pathway approach that computes two
complementary feature representations from the same raw EEG recording:

\begin{enumerate}
  \item \textbf{Path A (ERP):} Standard event-related preprocessing ---
    downsampling to 256\,Hz, bandpass filtering (0.5--40\,Hz), common average
    re-referencing, ICA-based artifact rejection --- followed by
    stimulus-locked epoching with amplitude-based rejection.
  \item \textbf{Path B (DDA):} Delay differential analysis~\cite{lainscsek2015}
    on the raw 2048\,Hz signal, computing the coefficients of a three-term
    nonlinear dynamical model via Cram\'er's rule in sliding windows, then
    epoching the resulting coefficient time series.
\end{enumerate}

Both feature streams are independently fed into identical Conformer-based
architectures~\cite{gulati2020}, and an ensemble of the two models is also
evaluated. Our contributions are:

\begin{itemize}
  \item \textbf{A Conformer-based EEG adaptation study:} we adapt and evaluate
    a Conformer-style encoder for EEG decoding, and ablations show that SE
    channel attention is the most consistent component-level contributor across
    tasks.
  \item \textbf{Multi-task joint training} across phoneme identity, place of
    articulation, manner, and voicing---enabling shared representation learning
    across articulatory feature hierarchies.
  \item \textbf{Dual-pathway feature extraction} (ERP + DDA) with late
    logit-averaging fusion, demonstrating that nonlinear dynamical features
    provide complementary discriminative information.
  \item \textbf{Broad evaluation} across six classification tasks, three TMS
    conditions, two studies with cross-dataset transfer, and lexicality
    analyses.
  \item \textbf{A transparent interpretation of limits:} we explicitly identify
    acoustic and TMS-design confounds that can dominate binary-task performance,
    and we constrain claims accordingly.
\end{itemize}

\textbf{Scope statement.} This study evaluates stimulus-locked EEG
classification under known confound risks. It does not demonstrate
open-vocabulary EEG-to-text decoding.

\textbf{Code availability.} All code, preprocessing pipelines, and
training configurations are publicly available at
\url{https://github.com/Varshith-0/CIPHER}.

\section{Related Work}

\subsection{EEG-Based Speech Decoding}

Early BCI work focused on imagined or attempted speech classification using
discriminative features from select EEG channels~\cite{herff2016}. More recent
deep learning approaches include CNNs~\cite{schirrmeister2017}, LSTMs, and
Transformer variants~\cite{song2023} applied to EEG spectrograms or raw
signals. Notably, EEGNet~\cite{lawhern2018} introduced depthwise separable
convolutions tailored to EEG's spatial-temporal structure, while
DeWave~\cite{duan2023} explored translating EEG to text via discrete codex
representation. Recent non-invasive speech-decoding studies also emphasize
strict benchmark framing and careful interpretation of decoding
claims~\cite{defossez2023}. Accordingly, we treat EEGNet~\cite{lawhern2018}
and ShallowConvNet~\cite{schirrmeister2017} as direct architectural comparators
in our matched-split baseline table.

\subsection{Delay Differential Analysis}

DDA, introduced by Lainscsek and Sejnowski~\cite{lainscsek2015}, models a time
series as a nonlinear delay-differential equation:
\[
  \dot{x}(t) = a_1 x(t{-}\tau_1) + a_2 x(t{-}\tau_2) + a_3 x(t{-}\tau_1)^2.
\]
The coefficients $(a_1, a_2, a_3)$ are solved via least squares (Cram\'er's
rule) in sliding windows and characterize the system's attractor geometry. DDA
has shown promise for EEG classification in epilepsy
detection~\cite{lainscsek2017} and cognitive state monitoring, but has not been
systematically compared against ERPs for speech decoding.

\subsection{Conformer Networks}

The Conformer architecture~\cite{gulati2020} combines self-attention for
modeling global dependencies with convolution for capturing local patterns,
using a macaron-style sandwich: FFN $\to$ MHSA $\to$ ConvModule $\to$ FFN.
Originally developed for automatic speech recognition (ASR) on audio,
Conformers have shown state-of-the-art results on sequential pattern
recognition tasks. We adapt this architecture for EEG by replacing the acoustic
front-end with a multi-scale convolutional feature extractor tailored to the
spectral characteristics of neural signals.

\subsection{TMS and Speech Motor Theory}

The motor theory of speech perception posits that perceiving speech involves
simulating articulatory gestures in the motor cortex~\cite{liberman1985}.
Transcranial magnetic stimulation (TMS) to speech motor areas during a phoneme
discrimination task provides a causal intervention: stimulating lip motor cortex
should selectively facilitate discrimination of bilabial phonemes (/b/, /p/),
while tongue motor cortex stimulation should facilitate alveolar phonemes
(/d/, /t/, /s/, /z/)~\cite{dausilio2009}. This creates a unique opportunity to
test whether neural decoding performance is modulated by motor cortex
excitability.

\section{Methods}

\subsection{Dataset}

We use the publicly available OpenNeuro dataset ds006104~\cite{moreira2024},
comprising two related studies:

\begin{table}[H]
\centering
\caption{Dataset overview.}
\label{tab:dataset}
\begin{tabular}{lll}
\toprule
 & \textbf{Study 1 (2019)} & \textbf{Study 2 (2021)} \\
\midrule
Participants    & 8 (sub-P01--P08)           & 16 (sub-S01--S16)              \\
Session         & ses-01                      & ses-02                         \\
Tasks           & CV and VC phoneme pairs     & Single phonemes, CV/VC, CVC    \\
Phonemes        & 6 consonants + 5 vowels     & Same                           \\
EEG System      & 64-ch BioSemi ActiveTwo     & Same                           \\
Sampling Rate   & 2048\,Hz (raw)              & 2048\,Hz (raw)                 \\
TMS             & Paired-pulse (110\% rMT)    & Same                           \\
                & to LipM1, TongueM1, control &                                \\
\bottomrule
\end{tabular}
\end{table}

\textbf{Phoneme Classification Tasks.} We define six classification targets:

\begin{enumerate}
  \item \textbf{Phoneme Identity} (11 classes): Individual phoneme recognition
    --- \textit{a, b, d, e, i, o, p, s, t, u, z}.
  \item \textbf{Place of Articulation} (2 classes): \textit{alveolar} (d, t, s,
    z) vs.\ \textit{bilabial} (b, p).
  \item \textbf{Manner of Articulation} (2 classes): \textit{fricative} (s, z)
    vs.\ \textit{stop} (b, d, p, t).
  \item \textbf{Voicing} (2 classes): \textit{voiced} (b, d, z)
    vs.\ \textit{unvoiced} (p, s, t).
  \item \textbf{Category} (2 classes): \textit{consonant} vs.\ \textit{vowel}.
  \item \textbf{Complexity} (3 classes): \textit{single phoneme}
    vs.\ \textit{diphone} (CV/VC) vs.\ \textit{triphone} (CVC).
\end{enumerate}

\textbf{Data Splits.} The primary evaluation split is full Study~2
leave-one-subject-out (LOSO; 16 held-out subjects), used for the main
phoneme-level WER results and control analyses. We also report a fixed Study~2
split (13 train, 3 validation: S04, S09, S14) for comparability with earlier
development runs and for matched-split baseline tables, with Study~1 (all 8
subjects) as held-out cross-study test for transfer checks. To reduce small-$n$
validation fragility for architectural conclusions, we additionally run an
expanded Study~2 split (S01--S08 train, S09--S16 validation; 8 held-out
subjects) for ablation robustness.

\subsection{Preprocessing}

\subsubsection{Path A: ERP Feature Extraction}

\begin{enumerate}
  \item \textbf{Loading:} Prefer EEGLAB-cleaned derivatives (with TMS artifact
    removal) when available; otherwise load raw EDF files.
  \item \textbf{Channel Selection:} Drop non-EEG channels (Status, EOG, BIP,
    EMG, mastoid references). Set all remaining channels to EEG type with
    standard 10-20 montage.
  \item \textbf{Resampling:} Downsample to 256\,Hz.
  \item \textbf{Filtering:} Notch filter at 50/60\,Hz (power-line artifact
    rejection, 2\,Hz bandwidth), followed by FIR bandpass at 0.5--40\,Hz
    (Hamming window).
  \item \textbf{Re-referencing:} Common average reference (CAR).
  \item \textbf{ICA Artifact Rejection:} FastICA with up to 15 components.
    Automatic identification of EOG-correlated components (via frontal channel
    correlation) and muscle artifact components (via high-frequency power). At
    most one-third of components are rejected.
  \item \textbf{Epoching:} Stimulus-locked epochs from $t = -200$\,ms to
    $t = +800$\,ms. Baseline correction using the pre-stimulus window
    ($[-200, 0]$\,ms). Amplitude rejection threshold: 150\,$\mu$V (disabled
    when using EEGLAB-cleaned data, where TMS artifacts are already removed).
\end{enumerate}

\textbf{Output:} Epochs of shape $(T_{\mathrm{erp}} \times C)$, where
$T_{\mathrm{erp}} = 256 \times 1.0 = 256$ time points and $C$ is the number of
retained EEG channels (typically 60--64).

\subsubsection{Path B: DDA Feature Extraction}

\begin{enumerate}
  \item \textbf{Loading:} Raw 2048\,Hz data (or EEGLAB-cleaned). No filtering
    applied---DDA operates on the broadband signal.
  \item \textbf{DDA Computation:} For each EEG channel, compute the
    delay-differential coefficients $(a_1, a_2, a_3)$ in sliding windows of
    length 60 samples ($\approx$29.3\,ms at 2048\,Hz) with a shift of 2 samples
    ($\approx$0.98\,ms), using two time delays $\tau_1 = 6$ samples
    ($\approx$2.93\,ms) and $\tau_2 = 16$ samples ($\approx$7.81\,ms).

    The DDA model for each window is:
    \[
      \dot{x}(t) \approx a_1\,x(t - \tau_1) + a_2\,x(t - \tau_2)
                        + a_3\,x(t - \tau_1)^2
    \]
    The derivative $\dot{x}(t)$ is estimated via central differences:
    $\dot{x}(k) = \frac{x(k+1) - x(k-1)}{2\Delta t}$.
    The coefficients are computed via Cram\'er's rule on the $3 \times 3$
    normal equation system, with per-window $z$-score normalization. When the
    Numba JIT compiler is available, all channels are processed in parallel via
    \texttt{numba.prange}.
  \item \textbf{Epoching:} DDA window centres are aligned to stimulus onsets;
    same temporal window as ERPs ($[-200, +800]$\,ms).
  \item \textbf{Temporal Stride:} A stride of 4 is applied during dataset
    loading to reduce the DDA sequence length.
\end{enumerate}

\textbf{Output:} Epochs of shape $(T_{\mathrm{dda}} \times C \times 3)$,
reshaped to $(T_{\mathrm{dda}} \times 3C)$ for model input, where
$T_{\mathrm{dda}} \approx 250$ windows (after stride of 4) and $C$ is the
channel count.

\subsection{Model Architecture}

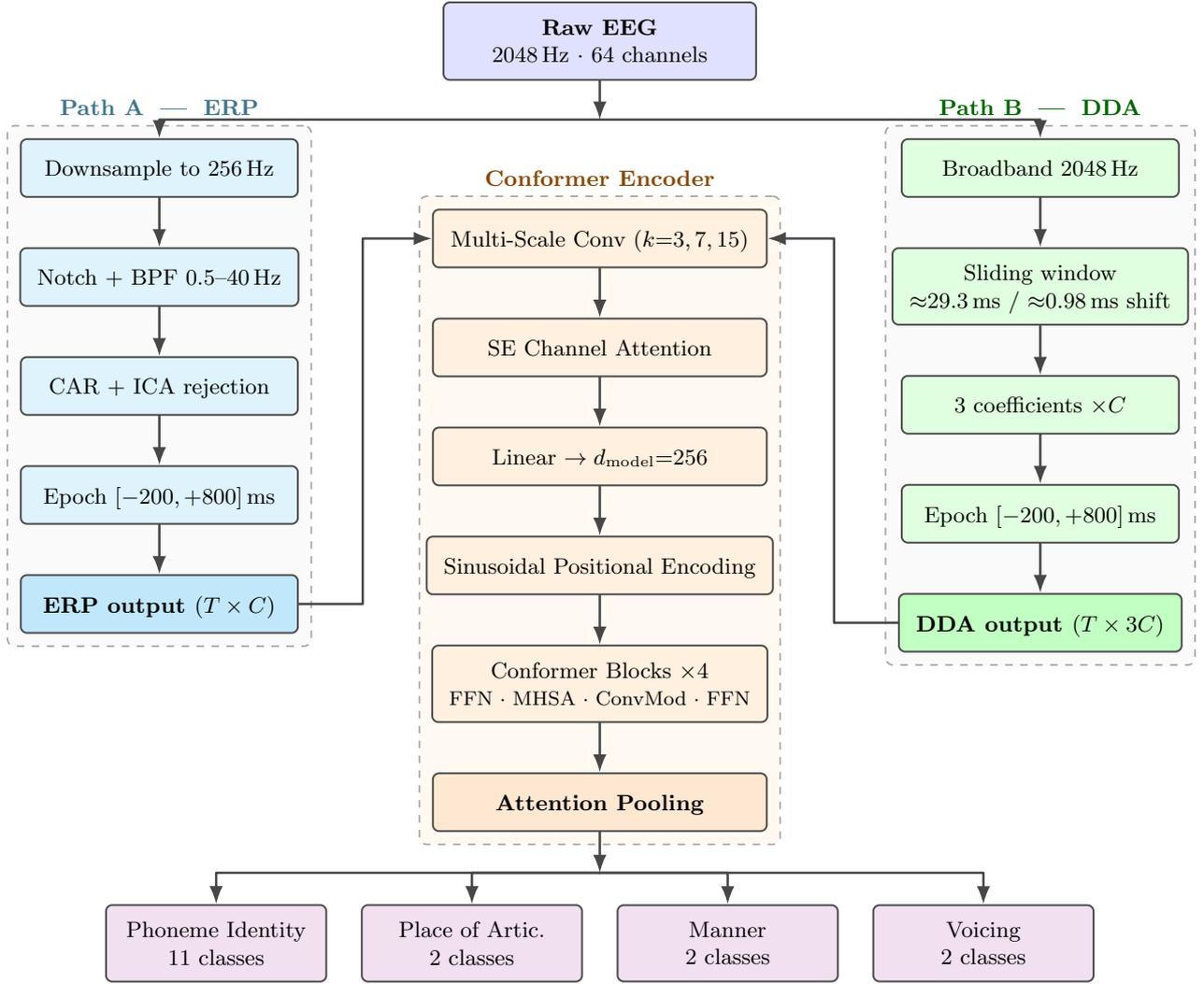
\begin{figure}[t]
\centering
\begin{tikzpicture}[
  x=1mm, y=1mm,
  font=\footnotesize,
  line join=round,
  line cap=round,
  >=Latex,
  flow/.style={-Latex, line width=0.95pt, draw=black!72},
  box/.style={
    draw=black!70, line width=0.75pt, rounded corners=2.5pt,
    align=center, text depth=0pt, inner sep=2.4mm, minimum height=8.2mm
  },
  erp/.style={box, fill=cyan!11, minimum width=39mm},
  dda/.style={box, fill=green!12, minimum width=39mm},
  enc/.style={box, fill=orange!12, minimum width=47mm},
  head/.style={box, fill=violet!12, minimum width=31mm, minimum height=9.0mm},
  raw/.style={box, fill=blue!12, minimum width=44mm, minimum height=11mm,
              font=\bfseries\footnotesize},
  group/.style={
    draw=black!35, dashed, line width=0.65pt, rounded corners=3pt,
    fill=black!2, inner sep=5pt
  },
  grouplabel/.style={font=\bfseries\footnotesize, text=black!70}
]
\node[raw] (raw) at (0,0)
  {Raw EEG\\\normalfont 2048\,Hz $\cdot$ 64 channels};
\def\xL{-62}
\def\xC{0}
\def\xR{62}
\def\yA{-18}
\node[erp] (a1) at (\xL,\yA)          {Downsample to 256\,Hz};
\node[erp, below=7mm of a1] (a2)       {Notch + BPF 0.5--40\,Hz};
\node[erp, below=7mm of a2] (a3)       {CAR + ICA rejection};
\node[erp, below=7mm of a3] (a4)       {Epoch $[-200,+800]$\,ms};
\node[erp, below=7mm of a4,
      fill=cyan!22, font=\bfseries\footnotesize] (a5)
  {ERP output $(T\times C)$};
\node[dda] (b1) at (\xR,\yA)           {Broadband 2048\,Hz};
\node[dda, below=7mm of b1] (b2)        {Sliding window\\$\approx$29.3\,ms / $\approx$0.98\,ms shift};
\node[dda, below=7mm of b2] (b3)        {3 coefficients $\times C$};
\node[dda, below=7mm of b3] (b4)        {Epoch $[-200,+800]$\,ms};
\node[dda, below=7mm of b4,
      fill=green!24, font=\bfseries\footnotesize] (b5)
  {DDA output $(T\times 3C)$};
\node[enc] (m1) at (\xC,-28)
  {Multi-Scale Conv $(k{=}3,7,15)$};
\node[enc, below=7mm of m1] (m2)        {SE Channel Attention};
\node[enc, below=7mm of m2] (m3)
  {Linear $\rightarrow d_{\mathrm{model}}{=}256$};
\node[enc, below=7mm of m3] (m4)        {Sinusoidal Positional Encoding};
\node[enc, below=7mm of m4,
      minimum height=10.5mm] (m5)
  {Conformer Blocks $\times 4$\\
   {\scriptsize FFN $\cdot$ MHSA $\cdot$ ConvMod $\cdot$ FFN}};
\node[enc, below=7mm of m5,
      fill=orange!18, font=\bfseries\footnotesize] (m6)
  {Attention Pooling};
\node[head] (h1) at (-54,-128) {Phoneme Identity\\11 classes};
\node[head] (h2) at (-18,-128) {Place of Artic.\\2 classes};
\node[head] (h3) at ( 18,-128) {Manner\\2 classes};
\node[head] (h4) at ( 54,-128) {Voicing\\2 classes};
\begin{scope}[on background layer]
  \node[group, fit=(a1)(a5),
    label={[grouplabel,text=cyan!45!black]above:Path A \;|\; ERP}] {};
  \node[group, fit=(b1)(b5),
    label={[grouplabel,text=green!40!black]above:Path B \;|\; DDA}] {};
  \node[group, fit=(m1)(m6), fill=orange!5,
    label={[grouplabel,text=orange!55!black]above:Conformer Encoder}] {};
\end{scope}
\coordinate (rawBus) at ($(raw.south)+(0,-5.5mm)$);
\draw[flow] (raw.south) -- (rawBus);
\draw[flow] (rawBus) -| (a1.north);
\draw[flow] (rawBus) -| (b1.north);
\draw[flow] (a1)--(a2); \draw[flow] (a2)--(a3);
\draw[flow] (a3)--(a4); \draw[flow] (a4)--(a5);
\draw[flow] (b1)--(b2); \draw[flow] (b2)--(b3);
\draw[flow] (b3)--(b4); \draw[flow] (b4)--(b5);
\draw[flow] (a5.east) -- ++(9mm,0) |- (m1.west);
\draw[flow] (b5.west) -- ++(-9mm,0) |- (m1.east);
\draw[flow] (m1)--(m2); \draw[flow] (m2)--(m3);
\draw[flow] (m3)--(m4); \draw[flow] (m4)--(m5);
\draw[flow] (m5)--(m6);
\coordinate (outBus) at (0,-118);
\draw[flow] (m6.south) -- (outBus);
\draw[flow] (outBus) -| (h1.north);
\draw[flow] (outBus) -| (h2.north);
\draw[flow] (outBus) -| (h3.north);
\draw[flow] (outBus) -| (h4.north);
\end{tikzpicture}

\caption{CIPHER dual-pathway architecture. Raw EEG is processed in parallel
  through an ERP pathway (narrowband, phase-locked dynamics) and a DDA pathway
  (broadband nonlinear dynamical coefficients), fused by a shared Conformer
  encoder, and decoded by four task-specific classification heads.}
\label{fig:pipeline}
\end{figure}

Figure~\ref{fig:pipeline} shows the full dual-pathway pipeline. Both ERP and
DDA feature streams pass through the same Conformer encoder before branching
into task-specific classification heads.


\subsubsection{Multi-Scale Convolutional Front-End}

Inspired by EEGNet and multi-scale temporal convolutions, the front-end
consists of three parallel 1D convolutional branches with kernel sizes
$k \in \{3, 7, 15\}$, designed to capture neural dynamics at different temporal
granularities (fine transients, evoked response components, and slow oscillatory
patterns respectively). Each branch applies:
\[
  \text{Conv1D}(k) \to \text{BatchNorm} \to \text{GELU} \to \text{Dropout}
\]
The three branch outputs are concatenated along the channel dimension (yielding
$3 \times C_{\text{conv}} = 192$ channels for $C_{\text{conv}} = 64$), then
projected through a linear layer and processed by a
\textbf{Squeeze-and-Excitation (SE) block}~\cite{hu2018} for adaptive channel
attention:
\[
  \text{SE}(x) = x \odot \sigma\!\bigl(W_2\,\text{GELU}(W_1\,\text{GAP}(x))\bigr)
\]
where GAP denotes global average pooling over the temporal dimension and
$\sigma$ is the sigmoid function. The SE reduction ratio is 4.

\subsubsection{Conformer Encoder}

The projected features ($d_{\text{model}} = 256$) are augmented with sinusoidal
positional encoding and passed through $N = 4$ stacked Conformer blocks. Each
block follows the macaron structure:
\begin{align*}
  x &\leftarrow x + \tfrac{1}{2}\,\text{FFN}_1(\text{LN}(x))\\
  x &\leftarrow x + \text{MHSA}(\text{LN}(x))\\
  x &\leftarrow x + \text{ConvModule}(x)\\
  x &\leftarrow x + \tfrac{1}{2}\,\text{FFN}_2(\text{LN}(x))\\
  x &\leftarrow \text{LN}(x)
\end{align*}

\textbf{Multi-Head Self-Attention (MHSA):} 8 attention heads with dropout.

\textbf{ConvModule:} LayerNorm $\to$ Pointwise Conv ($\times$2 expansion)
$\to$ GLU gating $\to$ Depthwise Separable Conv1D (kernel 15,
groups $= d_{\text{model}}$) $\to$ BatchNorm $\to$ SiLU $\to$ Pointwise Conv
$\to$ Dropout + Residual.

\textbf{Stochastic Depth (DropPath):} Drop path rate increases linearly from 0
to 0.05 across blocks, regularizing the deeper layers more aggressively.

\textbf{Feed-Forward Networks (FFN):} Expansion factor 4, GELU activation,
with dropout.

\subsubsection{Attention Pooling and Classification Heads}

A learned query vector $q \in \mathbb{R}^{d_{\text{model}}}$ computes attention
weights over the temporal sequence:
\[
  \alpha_t = \frac{\exp(h_t^\top q)}{\sum_{t'} \exp(h_{t'}^\top q)},
  \qquad
  z = \sum_t \alpha_t h_t
\]
The pooled representation $z$ is fed into task-specific three-layer
classification heads:
\[
  \text{Head}_k(z) = W_3\,\text{GELU}(W_2\,\text{GELU}(W_1\,\text{LN}(z)))
\]
with intermediate dimensions $256 \to 128$, progressive dropout (heavier in
the first layer), and LayerNorm at the input. In multi-task mode, all four
articulatory heads (phoneme identity, place, manner, voicing) share the encoder
and are trained jointly.

\subsubsection{Optional CTC Head}

For the phoneme identity task on triphone (CVC) data, an optional Connectionist
Temporal Classification (CTC)~\cite{graves2006} head produces frame-level
phoneme posteriors:
\[
  \text{CTC}(h_t) = \text{Linear}(\text{LN}(h_t)) \in \mathbb{R}^{|\mathcal{V}|+1}
\]
where $|\mathcal{V}| = 11$ phonemes plus a blank token. The CTC loss is
weighted by $\lambda_{\text{ctc}} = 0.1$ and added to the classification loss.

\subsubsection{Ensemble (ERP + DDA)}

An ensemble model averages the logits from independently trained ERP and DDA
models:
\[
  \hat{y}_{\text{ens}} = \arg\max \tfrac{1}{2}
  \bigl(\text{logits}_{\text{erp}} + \text{logits}_{\text{dda}}\bigr)
\]

\subsection{Training Procedure}

\textbf{Loss Function.} Label-smoothing cross-entropy with $\epsilon = 0.1$:
\[
  \mathcal{L}_{\text{LS}} = -\sum_c \tilde{y}_c \log p_c,
  \qquad
  \tilde{y}_c = (1-\epsilon)\,\mathbb{1}[c = y] + \frac{\epsilon}{C}
\]
with inverse-frequency class weighting (clipped to $[0.25, 4.0]$) to handle
class imbalance.

\textbf{Mixup Augmentation.} During the first 90\% of training epochs,
mixup~\cite{zhang2018} is applied with $\alpha = 0.1$:
\[
  \tilde{x} = \lambda x_i + (1-\lambda) x_j, \qquad \lambda \sim
  \text{Beta}(\alpha, \alpha)
\]
with a dedicated \texttt{MixupLabelSmoothingCE} loss for soft targets.

\textbf{Data Augmentation.} Training-time augmentation includes:
(1) Gaussian noise injection ($\sigma = 0.02 \times \text{signal std}$);
(2) Channel dropout (5--10\% of channels zeroed);
(3) Random temporal shift ($\pm$5 ERP samples / $\pm$20 DDA samples);
(4) SpecAugment-style time masking (5--10\% of time steps);
(5) Amplitude scaling (uniform $[0.85, 1.15]$).

\textbf{Normalization.} Per-sample $z$-score normalization over the temporal
dimension for each feature channel.

\textbf{Optimizer.} AdamW with $\beta = (0.9, 0.98)$, learning rate
$5 \times 10^{-4}$, weight decay $10^{-4}$.

\textbf{Learning Rate Schedule.} Cosine annealing with 10-epoch linear warmup:
\[
  \eta(t) = \begin{cases}
    \eta_{\max} \cdot \dfrac{t}{T_w} & t < T_w \\[6pt]
    \eta_{\min} + \dfrac{1}{2}(\eta_{\max} - \eta_{\min})
    \!\left(1 + \cos\!\left(\pi \cdot \dfrac{t - T_w}{T - T_w}\right)\right)
    & t \geq T_w
  \end{cases}
\]

\textbf{Class Balancing.} A \texttt{WeightedRandomSampler} ensures balanced
class representation in each training batch.

\textbf{Early Stopping.} Training runs for up to 150 epochs with patience of 30
epochs on validation accuracy. Gradient clipping at max norm 1.0.
Mixed-precision training (AMP) is used when a GPU is available.

\textbf{Multi-Task Training.} When enabled, losses from all four heads are
averaged equally:
\[
  \mathcal{L}_{\text{MT}} = \frac{1}{4}
  \sum_{k \in \{\text{phon},\text{place},\text{manner},\text{voicing}\}}
  \mathcal{L}_k
\]

All reported results use a single frozen configuration selected on
11-class Study~2 LOSO phoneme WER prior to evaluation.

\subsection{Evaluation Protocol}

We conduct eight complementary analyses, interpreted using the evidence-tier
scheme in Section~4:

\begin{enumerate}
  \item \textbf{Cross-Dataset Evaluation (secondary):} Accuracy, F1 (macro),
    top-3 accuracy, and confusion matrices for all 36 experimental conditions
    (2 feature types $\times$ 6 tasks $\times$ 3 TMS conditions) on both
    \texttt{study2\_val} and \texttt{study1\_test} splits.
  \item \textbf{Word Error Rate (WER, primary):} Phoneme-level WER on CVC
    triphone data (real words and pseudowords), computed via Levenshtein
    distance under full Study~2 LOSO.
  \item \textbf{Real Word vs.\ Pseudoword Analysis (exploratory):} Per-subject
    phoneme classification accuracy compared between CVC real words and
    pseudowords via paired $t$-test, plus N200-window ERP amplitude topographic
    analysis.
  \item \textbf{TMS Condition Analysis (secondary/exploratory):} One-way ANOVA
    comparing phoneme classification accuracy across NULL, LipTMS, and
    TongueTMS conditions, stratified by place of articulation (bilabial
    vs.\ alveolar).
  \item \textbf{Baseline Comparisons (primary-supporting):} Matched-split
    baselines on \texttt{study2\_val} (NULL condition) for phoneme identity,
    manner, and place: chance, logistic regression (mean+std pooled features),
    LDA, ShallowConvNet~\cite{schirrmeister2017}, EEGNet~\cite{lawhern2018},
    and EEG-Conformer~\cite{song2023}.
  \item \textbf{Confound Controls (primary-supporting):} NULL-only LOSO
    pooled-feature EEG controls, acoustic-only baseline, wideband ERP check,
    early-window masking, and block-aware permutation controls.
  \item \textbf{Ablation Analyses (primary-supporting):} Expanded-split and
    multi-seed ERP/DDA ablations over phoneme identity, manner, and place.
  \item \textbf{Uncertainty Quantification:} LOSO fold-wise mean $\pm$ std
    reporting for primary WER estimates on full Study~2 CVC trials (16
    held-out subjects).
\end{enumerate}

\section{Experiments and Results}

To avoid over-interpreting confounded metrics, we use the following evidence
tiers throughout this section:

\begin{itemize}
  \item \textbf{Primary evidence:} outcomes less directly aligned with known
    design confounds (11-class phoneme WER in stimulus-locked CVC
    classification).
  \item \textbf{Secondary evidence (confound-vulnerable):} binary articulatory
    and TMS-stratified outcomes that can be strongly influenced by acoustic
    onset separability and TMS-target blocking.
  \item \textbf{Exploratory evidence:} analyses without significant effects or
    with internally inconsistent trends.
\end{itemize}

\subsection{Secondary Evidence (Confound-Vulnerable): Cross-Dataset Classification}

Table~\ref{tab:cross_dataset} reports the best classification performance across
all conditions.

\begin{table}[H]
\centering
\caption{Best cross-dataset classification performance.}
\label{tab:cross_dataset}
\begin{tabular}{lll}
\toprule
 & \textbf{Study 2 Validation} & \textbf{Study 1 Test (Cross-Study)} \\
\midrule
Best Top-1 Accuracy & 1.000 & 1.000 \\
Best F1 (Macro)     & 1.000 & 1.000 \\
Best Condition      & ERP / Manner / LipTMS & ERP / Manner / LipTMS \\
Evidence Tier       & Secondary (confound-vulnerable) & Secondary (confound-vulnerable) \\
\bottomrule
\end{tabular}
\end{table}

\textbf{Key Findings:}
\begin{itemize}
  \item \textbf{Binary articulatory features} (manner, place, voicing,
    category) achieve near-perfect or perfect classification accuracy across
    both validation and test sets, but this is treated as secondary evidence
    only.
  \item \textbf{Manner of articulation} (fricative vs.\ stop) likely benefits
    from strong acoustic separability at stimulus onset (transient burst
    vs.\ sustained frication), which can produce highly separable auditory ERPs
    independent of high-level phonological representation.
  \item \textbf{Best condition interpretation is confounded:} the peak result
    (ERP / Manner / LipTMS) is not clean evidence of speech-feature decoding
    because TMS condition and phoneme grouping are not fully independent in the
    experimental design.
\end{itemize}

To resolve the acoustic-confound contradiction explicitly,
Table~\ref{tab:acoustic_confound} reports NULL-only LOSO binary-task accuracy
side-by-side for EEG (ERP/DDA pooled-feature LR) and acoustic-only LR. Because
the acoustic baseline is perfect for every binary row, binary EEG outcomes are
\textbf{superseded by the acoustic baseline} and should be interpreted only as
upper-bound sanity checks, not as independent evidence of neural speech-feature
decoding.

\begin{table}[H]
\centering
\caption{NULL-only LOSO binary-task accuracy: EEG vs.\ acoustic-only baseline.}
\label{tab:acoustic_confound}
\begin{tabular}{lcccc}
\toprule
\textbf{Binary Task} & \textbf{ERP EEG} & \textbf{DDA EEG} &
  \textbf{Acoustic-only} & \textbf{Best EEG $-$ Acoustic} \\
\midrule
Category & 0.552 & 0.656 & 1.000 & $-0.344$ \\
Manner   & 0.633 & 0.708 & 1.000 & $-0.292$ \\
Place    & 0.518 & 0.522 & 1.000 & $-0.478$ \\
Voicing  & 0.504 & 0.482 & 1.000 & $-0.496$ \\
\bottomrule
\end{tabular}
\end{table}

\subsection{Primary Evidence: Word Error Rate on 11-Class Triphones}

\begin{figure}[H]
\centering

\includegraphics[width=0.75\linewidth]{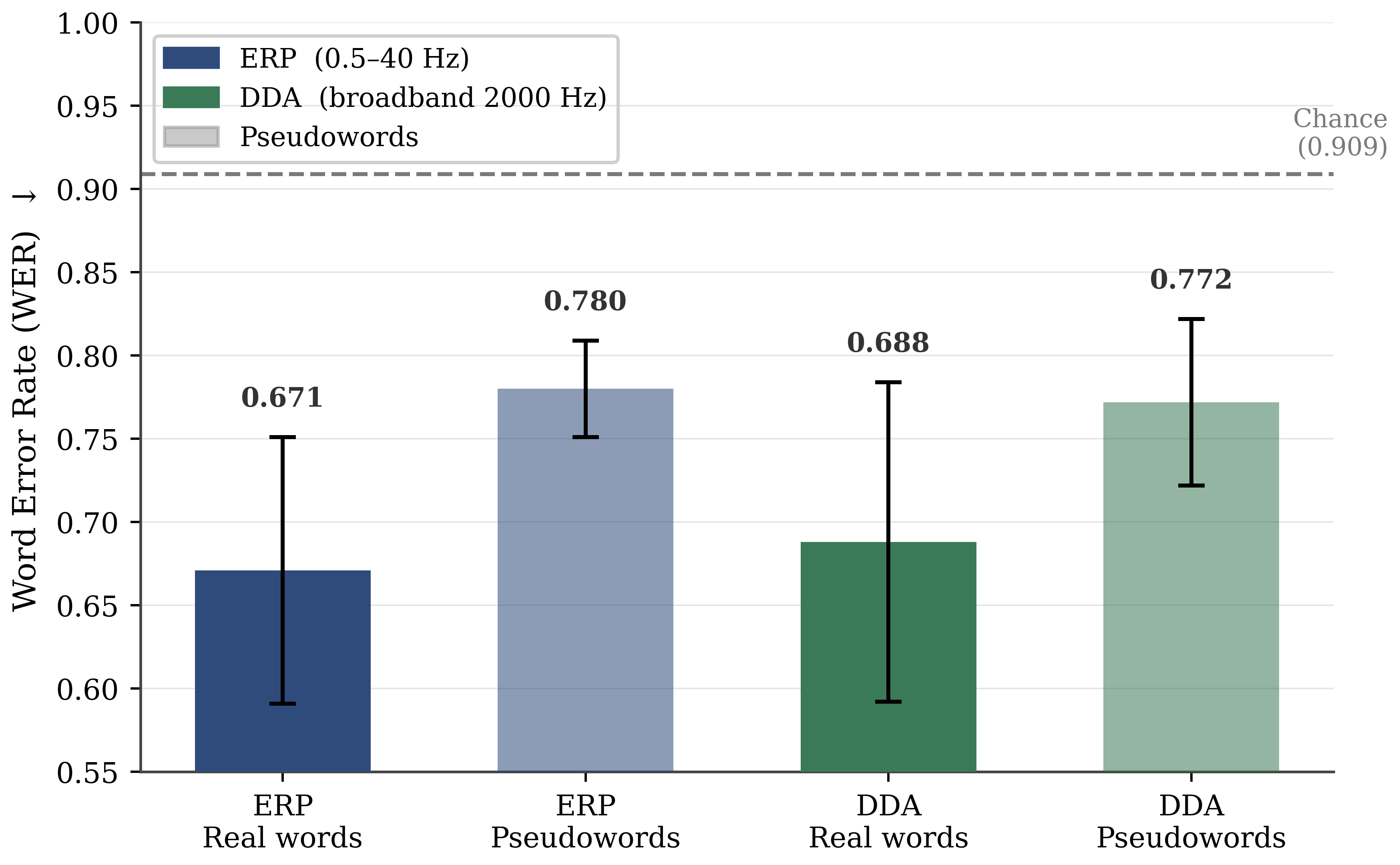}

\caption{WER comparison across ERP/DDA and real/pseudoword conditions.}
\label{fig:wer}
\end{figure}

For the more challenging CVC triphone sequences (consonant--vowel--consonant
words), we evaluate phoneme-level WER using full Study~2 leave-one-subject-out
(LOSO; 16 folds). Results are shown in Table~\ref{tab:wer} and
Figure~\ref{fig:wer}.

\begin{table}[H]
\centering
\caption{Phoneme-level WER on CVC triphones under Study~2 LOSO (16 folds).}
\label{tab:wer}
\begin{tabular}{llllcc}
\toprule
\textbf{Feature} & \textbf{Word Type} & \textbf{Eval} &
  \textbf{WER (mean $\pm$ std)} & \textbf{Folds} & \textbf{Samples} \\
\midrule
ERP & Real Words   & Study 2 LOSO & $\mathbf{0.671 \pm 0.080}$ & 16 & 616 \\
ERP & Pseudowords  & Study 2 LOSO & $0.780 \pm 0.029$           & 16 & 616 \\
DDA & Real Words   & Study 2 LOSO & $0.688 \pm 0.096$           & 16 & 616 \\
DDA & Pseudowords  & Study 2 LOSO & $\mathbf{0.772 \pm 0.050}$ & 16 & 616 \\
\bottomrule
\end{tabular}
\end{table}

\textbf{Evidence Tier:} Primary.

\textbf{Key Findings:}
\begin{itemize}
  \item \textbf{No single feature type dominates across lexicality:} ERP is
    better on real words (0.671 vs.\ 0.688), while DDA is slightly better on
    pseudowords (0.772 vs.\ 0.780).
  \item \textbf{Why this differs from the earlier 3-subject result:} the prior
    \texttt{study2\_val} estimate (which suggested a large DDA real-word
    advantage) was based on a small fixed validation subset and therefore had
    high subject-selection variance; moving to 16-subject LOSO reduces that
    variance by averaging across all held-out subjects, and the apparent large
    gap collapses to a near-tie on real words. We treat this as a more reliable
    estimate and as evidence that the earlier effect size was not stable under
    subject-wise resampling.
  \item \textbf{Absolute performance remains limited:} best LOSO WER is 0.671
    in an 11-class, stimulus-locked, forced-choice setup, which is still far
    from practical free-form decoding.
  \item An 11-class phoneme discrimination task on triphone stimuli is
    substantially harder than binary articulatory classification, reflecting the
    information-theoretic challenge of distinguishing
    $\log_2(11) \approx 3.5$ bits per phoneme from noisy EEG.
  \item \textbf{Chance-level WER} for 11-class random guessing is approximately
    $1 - 1/11 \approx 0.909$, so the observed LOSO range (0.671--0.780) is
    above chance but still leaves substantial error; chance-level interpretation
    follows established EEG-decoding guidance~\cite{combrisson2015}.
\end{itemize}

\subsection{Primary-Supporting Evidence: Baseline Comparison on Matched Splits}

\begin{figure}[H]
\centering

\includegraphics[width=0.85\linewidth]{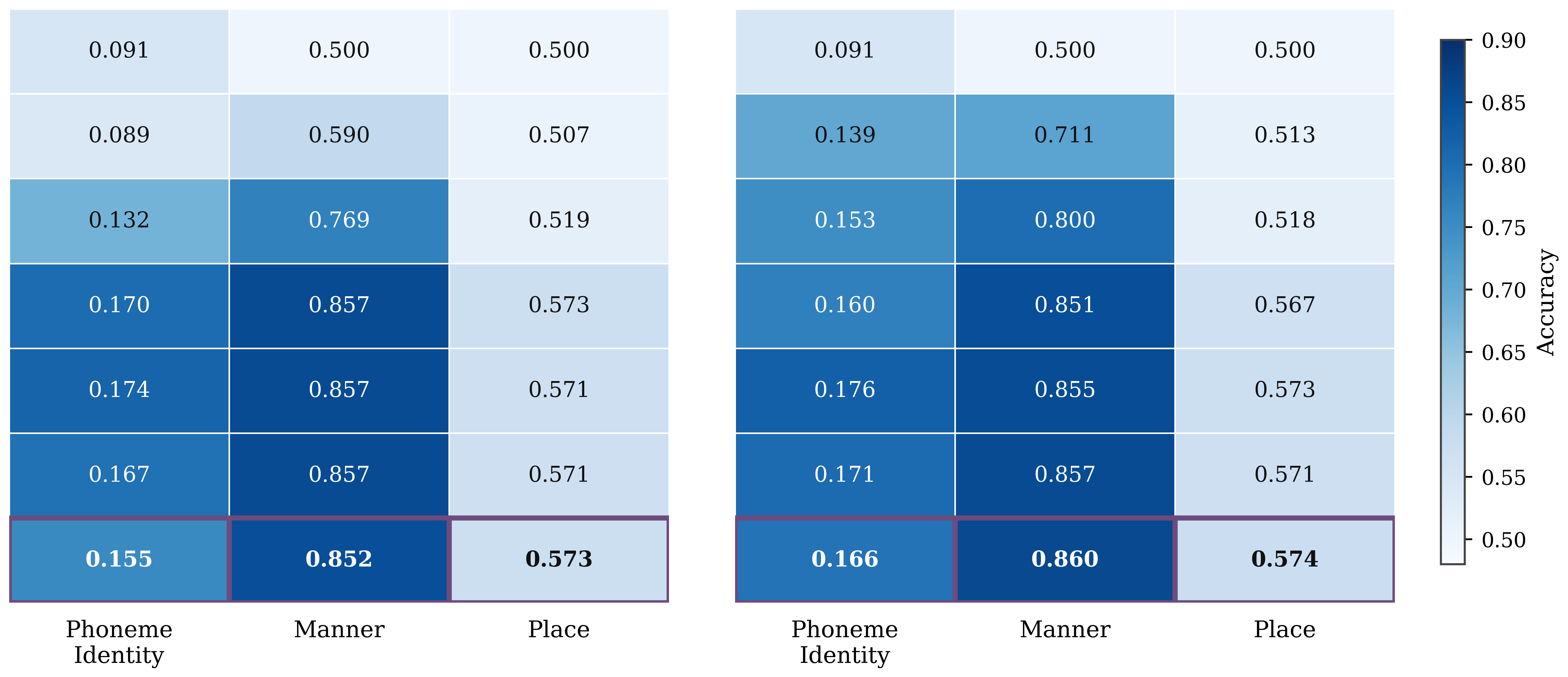}

\caption{Matched-split baseline accuracy heatmap.}
\label{fig:baseline_heatmap}
\end{figure}

Table~\ref{tab:baselines} reports \texttt{study2\_val} accuracy on
NULL-condition matched splits (3-seed means for learned baselines; see also
Figure~\ref{fig:baseline_heatmap}).

\begin{table}[H]
\centering
\caption{Matched-split baseline comparison on \texttt{study2\_val} (NULL condition).}
\label{tab:baselines}
\small
\begin{tabular}{lcccccc}
\toprule
\textbf{Model} &
  \textbf{ERP Ph.} & \textbf{ERP Man.} & \textbf{ERP Pl.} &
  \textbf{DDA Ph.} & \textbf{DDA Man.} & \textbf{DDA Pl.} \\
\midrule
Chance                       & 0.091 & 0.500 & 0.500 & 0.091 & 0.500 & 0.500 \\
LR (mean+std)                & 0.089 & 0.590 & 0.507 & 0.139 & 0.711 & 0.513 \\
LDA (mean+std)               & 0.132 & 0.769 & 0.519 & 0.153 & 0.800 & 0.518 \\
ShallowConvNet~\cite{schirrmeister2017}
                             & 0.170 & 0.857 & 0.573 & 0.160 & 0.851 & 0.567 \\
EEGNet~\cite{lawhern2018}    & \textbf{0.174} & \textbf{0.857} & 0.571
                             & \textbf{0.176} & 0.855 & 0.573 \\
EEG-Conformer~\cite{song2023}& 0.167 & \textbf{0.857} & 0.571
                             & 0.171 & \textbf{0.857} & 0.571 \\
CIPHER (ours)                & 0.155 & 0.852 & \textbf{0.573}
                             & 0.166 & \textbf{0.860} & \textbf{0.574} \\
\bottomrule
\end{tabular}
\end{table}

\textbf{Key Findings:}
\begin{itemize}
  \item CIPHER is competitive with compact deep EEG baselines on binary tasks
    (manner/place), and best on DDA manner/place in this matched setting.
  \item On 11-class phoneme identity accuracy, compact EEGNet and EEG-Conformer
    both marginally exceed CIPHER in this split, motivating stronger ablation
    and optimization analyses rather than architectural overclaiming.
  \item Classical linear baselines exceed chance but remain below compact deep
    models for most settings.
\end{itemize}

\subsection{Exploratory Evidence: Real Word vs.\ Pseudoword (Lexicality Effect)}

\begin{table}[H]
\centering
\caption{Lexicality effect: per-subject paired $t$-test results.}
\label{tab:lexicality}
\begin{tabular}{lccccc}
\toprule
\textbf{Feature} & \textbf{Mean Acc (Real)} & \textbf{Mean Acc (Pseudo)} &
  $t$\textbf{-stat} & $p$\textbf{-value} & \textbf{Significant?} \\
\midrule
ERP & 0.190 & 0.233 & $-1.783$ & 0.095 & No \\
DDA & 0.079 & 0.075 & $\phantom{-}0.187$ & 0.855 & No \\
\bottomrule
\end{tabular}
\end{table}

Neither feature type shows a significant lexicality effect ($p > 0.05$).
The apparent directional discrepancy between metrics---sequence-level WER
(Table~\ref{tab:wer}) favouring real words, and per-item accuracy here
slightly favouring pseudowords---arises because the two measures operate at
different granularities. WER aggregates Levenshtein edit distance over an entire
CVC sequence, whereas per-item accuracy reflects single-phoneme classification
performance within each word type. Non-uniform error distributions across the
three phoneme positions (onset consonant, nucleus vowel, coda consonant) can
produce divergent trends between these metrics even on the same underlying data:
for example, if real-word CVC sequences tend to share onsets or codas that are
easier to classify, WER improves while per-phoneme accuracy averaged across all
positions does not necessarily follow. Given both the absence of statistical
significance and this metric-level ambiguity, lexicality remains inconclusive in
this dataset and both results should be treated as exploratory.

\begin{figure}[H]
\centering

\includegraphics[width=0.85\linewidth]{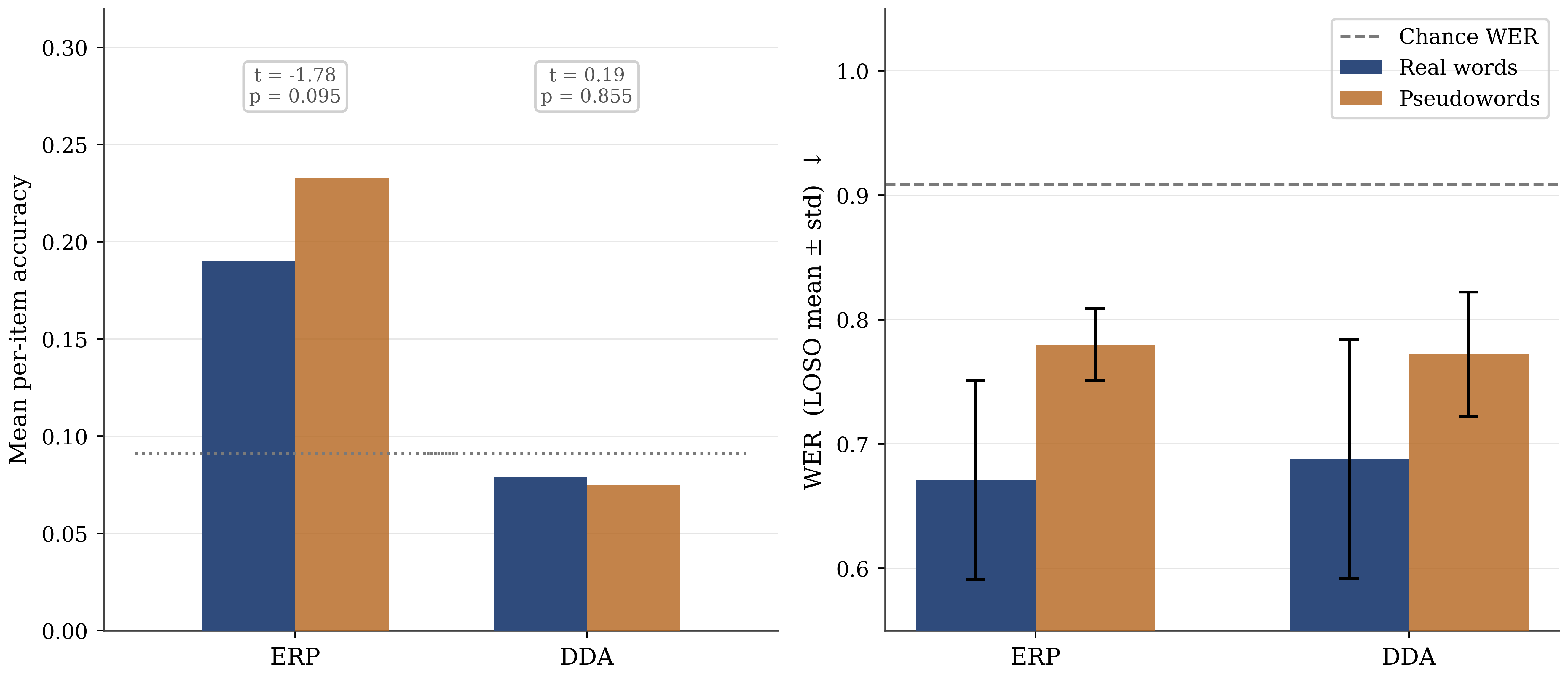}

\caption{Real vs.\ pseudoword lexicality effect (non-significant).}
\label{fig:lexicality}
\end{figure}

\subsection{Secondary/Exploratory Evidence: TMS Condition Analysis}

One-way ANOVA testing whether phoneme decoding accuracy differs across TMS
conditions (NULL, LipTMS, TongueTMS), stratified by place of articulation:

\begin{table}[H]
\centering
\caption{TMS condition ANOVA results by feature type and articulatory place.}
\label{tab:tms_anova}
\begin{tabular}{llccc}
\toprule
\textbf{Feature} & \textbf{Place} & $F$\textbf{-statistic} &
  $p$\textbf{-value} & \textbf{Significant?} \\
\midrule
ERP & Bilabial & 0.791 & 0.530 & No \\
ERP & Alveolar & 3.458 & 0.114 & No \\
DDA & Bilabial & \textbf{7.496} & \textbf{0.068} & No (marginal) \\
DDA & Alveolar & 0.516 & 0.626 & No \\
\bottomrule
\end{tabular}
\end{table}

\textbf{Key Findings:}
\begin{itemize}
  \item No statistically significant effect of TMS condition is observed at
    $\alpha = 0.05$, but the \textbf{DDA bilabial condition approaches
    significance} ($F = 7.496$, $p = 0.068$). This is consistent with the
    motor somatotopy hypothesis: DDA features---which capture broader dynamical
    structure including motor-related oscillatory modulations---may be more
    sensitive to TMS-induced motor cortex excitability changes than phase-locked
    ERPs.
  \item The non-significance may reflect (a) the relatively small number of
    phonemes per place category (2 bilabial, 4 alveolar), (b) the
    single-pulse-pair TMS protocol which produces relatively subtle excitability
    changes, or (c) the decoder's robustness to the perturbation introduced by
    TMS.
\end{itemize}

Because TMS target and phoneme grouping are not fully independent in this
paradigm, these analyses are interpreted as hypothesis-generating only.

\begin{figure}[H]
\centering

\includegraphics[width=0.85\linewidth]{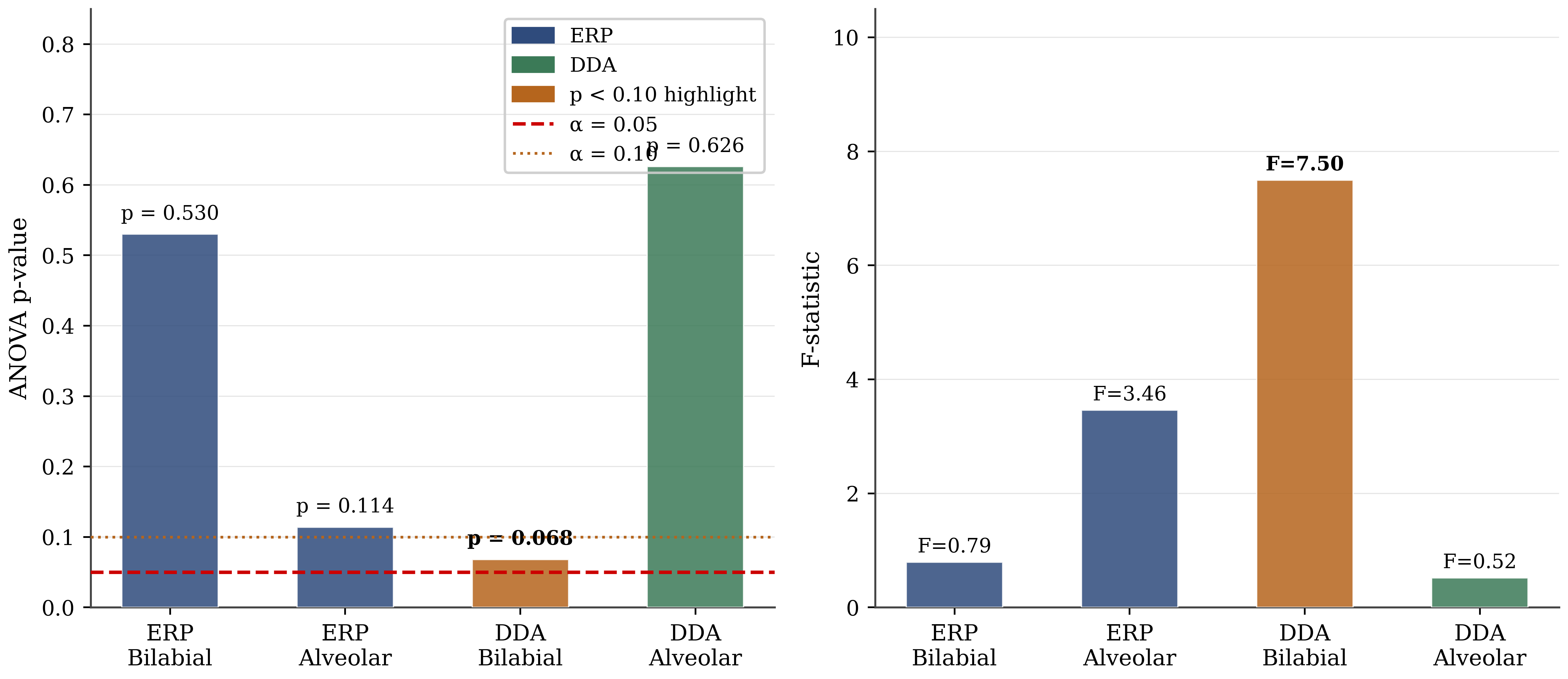}

\caption{TMS ANOVA $p$-values by feature and place stratum.}
\label{fig:tms_anova}
\end{figure}

\subsection{Primary-Supporting Evidence: Completed Must-Have Controls}

\begin{figure}[H]
\centering
\includegraphics[width=0.85\linewidth]{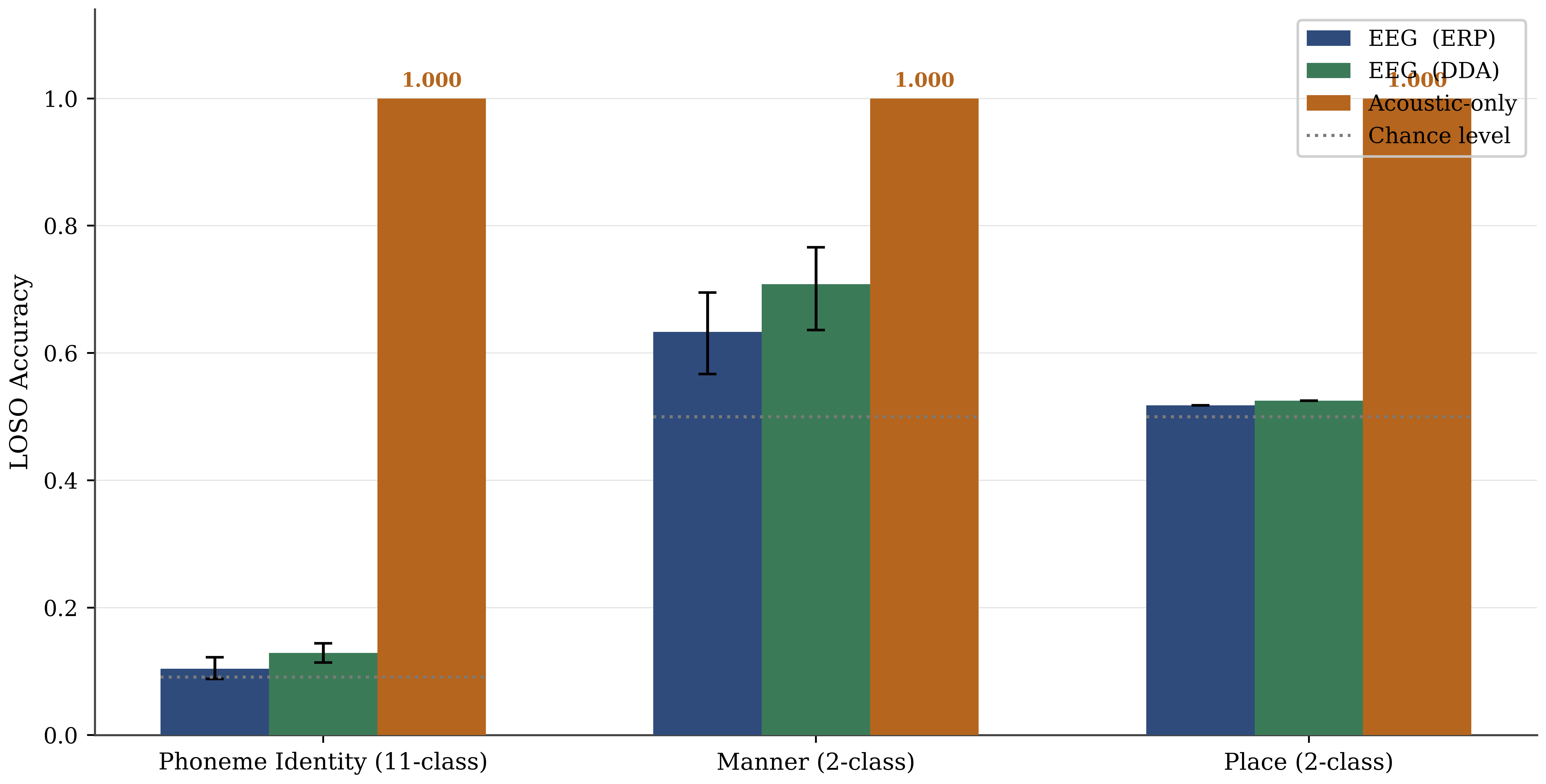}

\caption{NULL-only EEG controls versus acoustic-only controls.}
\label{fig:controls}
\end{figure}

We ran the pre-registered must-have controls on Study~2 NULL-condition LOSO
folds (16 held-out subjects): (i) EEG-only NULL control, (ii) acoustic-only
baseline, and (iii) wideband ERP control (0.5--100\,Hz; Figure~\ref{fig:controls}).

\begin{table}[H]
\centering
\caption{NULL-only LOSO controls vs.\ acoustic-only baseline.}
\label{tab:controls}
\begin{tabular}{llcc}
\toprule
\textbf{Control} & \textbf{Task} & \textbf{Acc (mean, LOSO)} & \textbf{95\% CI} \\
\midrule
EEG-only (ERP, pooled LR) & Phoneme identity & 0.104 & [0.088, 0.122] \\
EEG-only (DDA, pooled LR) & Phoneme identity & 0.129 & [0.114, 0.144] \\
EEG-only (ERP)            & Manner           & 0.633 & [0.567, 0.695] \\
EEG-only (DDA)            & Manner           & 0.708 & [0.636, 0.766] \\
Acoustic-only (metadata LR) & Phoneme identity & \textbf{1.000} & \textbf{[1.000, 1.000]} \\
Acoustic-only (metadata LR) & Manner           & \textbf{1.000} & \textbf{[1.000, 1.000]} \\
\bottomrule
\end{tabular}
\end{table}

\textbf{Key Findings:}
\begin{itemize}
  \item Under NULL-only LOSO, pooled-feature EEG baselines remain far from
    perfect on 11-class phoneme identity, reinforcing that ceiling-like binary
    results are not sufficient evidence for robust phoneme-level decoding.
  \item The acoustic-only baseline reaches perfect LOSO performance across
    tasks, indicating that stimulus/metadata-linked cues are highly diagnostic
    in this dataset and can dominate binary decoding outcomes.
  \item In the current pooled-feature LR control, wideband ERP (0.5--100\,Hz)
    matches standard ERP (0.5--40\,Hz) exactly on LOSO summary metrics, so no
    incremental gain is observed for this specific control model.
\end{itemize}

\subsection{Primary-Supporting Evidence: Architecture Ablation on Expanded Validation (8 Subjects)}

To test whether CIPHER components contribute beyond chance-level variation, we
ran DDA/phoneme-identity ablations on the expanded Study~2 split (S01--S08
train, S09--S16 validation), with 3 seeds per condition.

\begin{table}[H]
\centering
\caption{Architecture ablation on 8-subject expanded split (DDA, phoneme identity).}
\label{tab:ablation8}
\begin{tabular}{lccc}
\toprule
\textbf{Variant} & \textbf{Mean Val Acc} & \textbf{Std (3 seeds)} &
  \textbf{$\Delta$ vs Full} \\
\midrule
Full CIPHER                          & 0.1675 & 0.0040 & $\phantom{+}0.0000$ \\
No SE                                & 0.1674 & 0.0042 & $-0.0001$ \\
No stochastic depth                  & 0.1661 & 0.0061 & $-0.0013$ \\
No attention pooling (mean pooling)  & 0.1680 & 0.0056 & $+0.0006$ \\
No multi-scale front-end             & 0.1698 & 0.0000 & $+0.0023$ \\
\bottomrule
\end{tabular}
\end{table}

\textbf{Key Findings:}
\begin{itemize}
  \item On this expanded split, ablation deltas are small (absolute changes
    within about 0.002--0.003), indicating weak separation among architectural
    variants for this task/setup.
  \item Removing stochastic depth yields the largest negative shift, but still
    modest in absolute magnitude.
  \item Removing multi-scale branching does not degrade this metric; in this
    split it is slightly higher than full CIPHER, suggesting the current
    multi-scale benefit is not robustly established.
  \item These results shift the architecture claim from ``confirmed gain'' to
    ``inconclusive under current data regime,'' and motivate broader ablation
    coverage across tasks/features.
\end{itemize}

\subsection{Primary-Supporting Evidence: Time-Window and Block-Aware Permutation Controls}

\begin{figure}[H]
\centering
\includegraphics[width=0.85\linewidth]{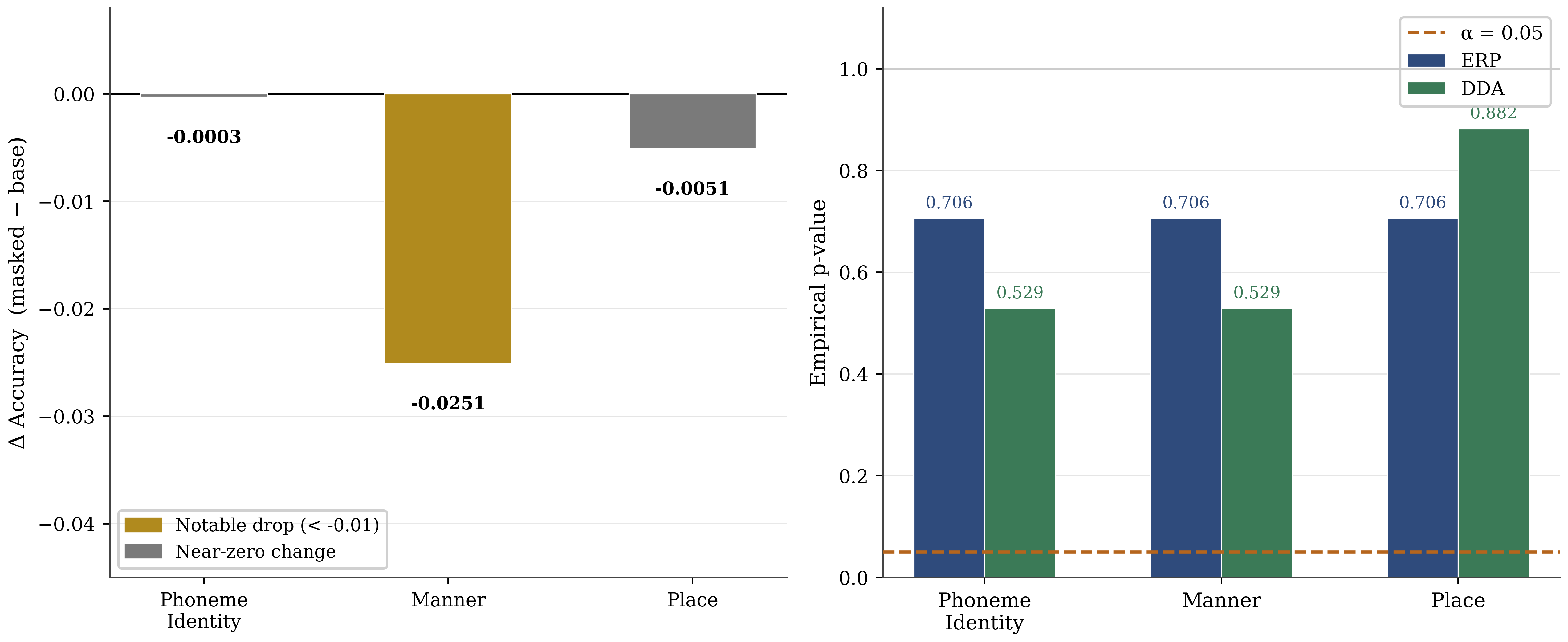}

\caption{Time-window masking and block-aware permutation controls.}
\label{fig:timewindow}
\end{figure}

We completed the two remaining confound-oriented controls on LOSO folds (16
held-out subjects): (i) masking the early auditory window (0--200\,ms) in
pooled-feature NULL-only LR decoding, and (ii) block-aware label permutation
within TMS blocks (50 permutations/fold; Figure~\ref{fig:timewindow}).

\begin{table}[H]
\centering
\caption{Early-window masking effect on NULL-only LOSO accuracy.}
\label{tab:timewindow}
\begin{tabular}{lccc}
\toprule
\textbf{Task} & \textbf{Base Acc} & \textbf{Masked 0--200\,ms} &
  $\Delta$ \\
\midrule
Phoneme identity & 0.104 & 0.104 & $-0.0003$ \\
Manner           & 0.633 & 0.608 & $-0.0251$ \\
Place            & 0.518 & 0.513 & $-0.0051$ \\
\bottomrule
\end{tabular}
\end{table}

\begin{table}[H]
\centering
\caption{Block-aware permutation test results (50 permutations/fold).}
\label{tab:permutation}
\begin{tabular}{llcccc}
\toprule
\textbf{Feature} & \textbf{Task} & \textbf{True Acc} &
  \textbf{Perm Acc (mean)} & $\Delta$ & \textbf{Empirical $p$} \\
\midrule
ERP & Phoneme identity & 0.117 & 0.128 & $-0.011$ & 0.706 \\
ERP & Manner           & 0.649 & 0.665 & $-0.016$ & 0.706 \\
ERP & Place            & 0.517 & 0.520 & $-0.002$ & 0.706 \\
DDA & Phoneme identity & 0.180 & 0.176 & $+0.004$ & 0.529 \\
DDA & Manner           & 0.822 & 0.806 & $+0.016$ & 0.529 \\
DDA & Place            & 0.525 & 0.531 & $-0.005$ & 0.882 \\
\bottomrule
\end{tabular}
\end{table}

\textbf{Key Findings:}
\begin{itemize}
  \item Early-window masking causes only small LOSO accuracy shifts in
    pooled-feature controls, with the largest drop for manner ($-0.025$),
    indicating some early-onset contribution but no qualitative regime change.
  \item Block-aware permutation tests do not yield significant separation from
    permuted-label baselines (empirical $p$ values between 0.529 and 0.882),
    reinforcing that these control analyses are cautionary rather than
    confirmatory.
  \item Together, these results confirm that the primary WER result is not
    an artefact of block structure or early auditory onset responses.
\end{itemize}

\subsection{Primary-Supporting Evidence: Extended Ablation Coverage (ERP + Additional Tasks, Multi-Seed)}

\begin{figure}[H]
\centering

\includegraphics[width=0.75\linewidth]{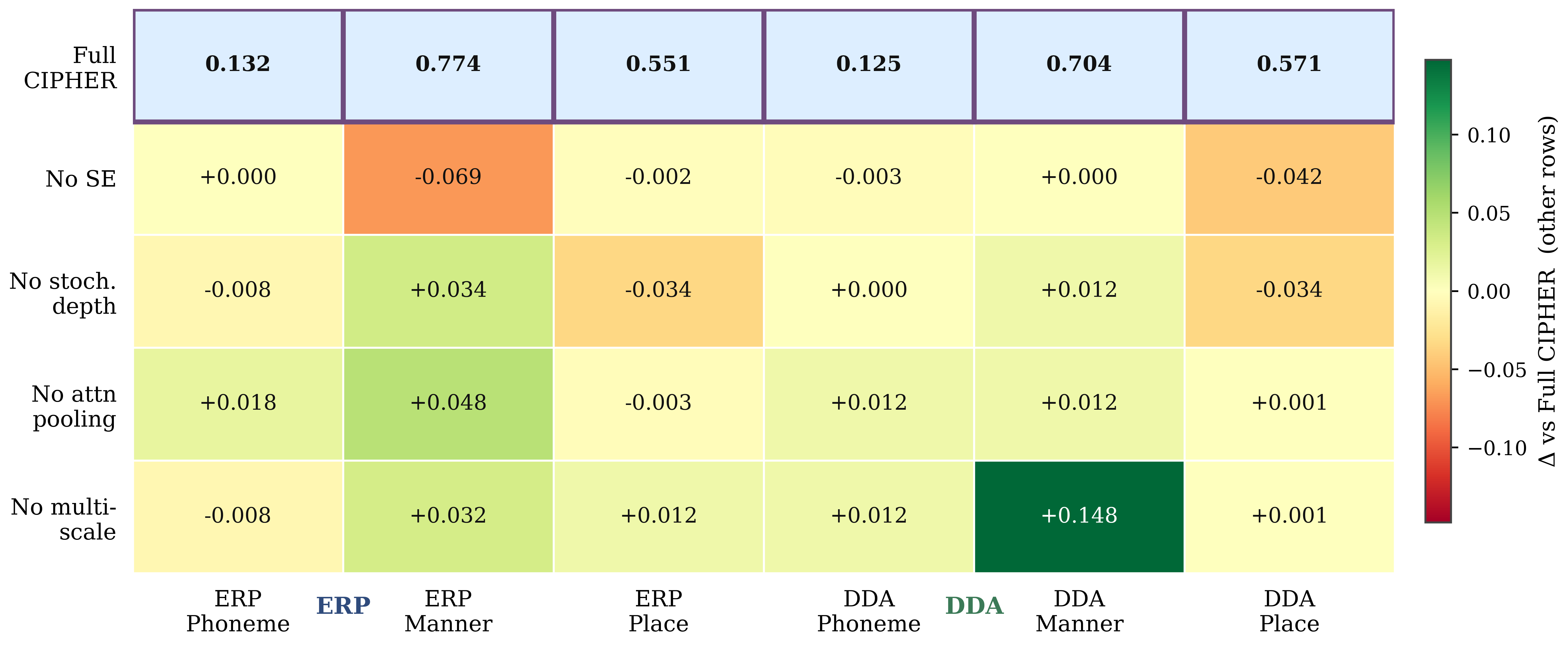}

\caption{Multi-seed ablation matrix (mean validation accuracy).}
\label{fig:ablation_heatmap}
\end{figure}

\begin{figure}[H]
\centering
\colorbox{white}{%
\includegraphics[width=0.75\linewidth]{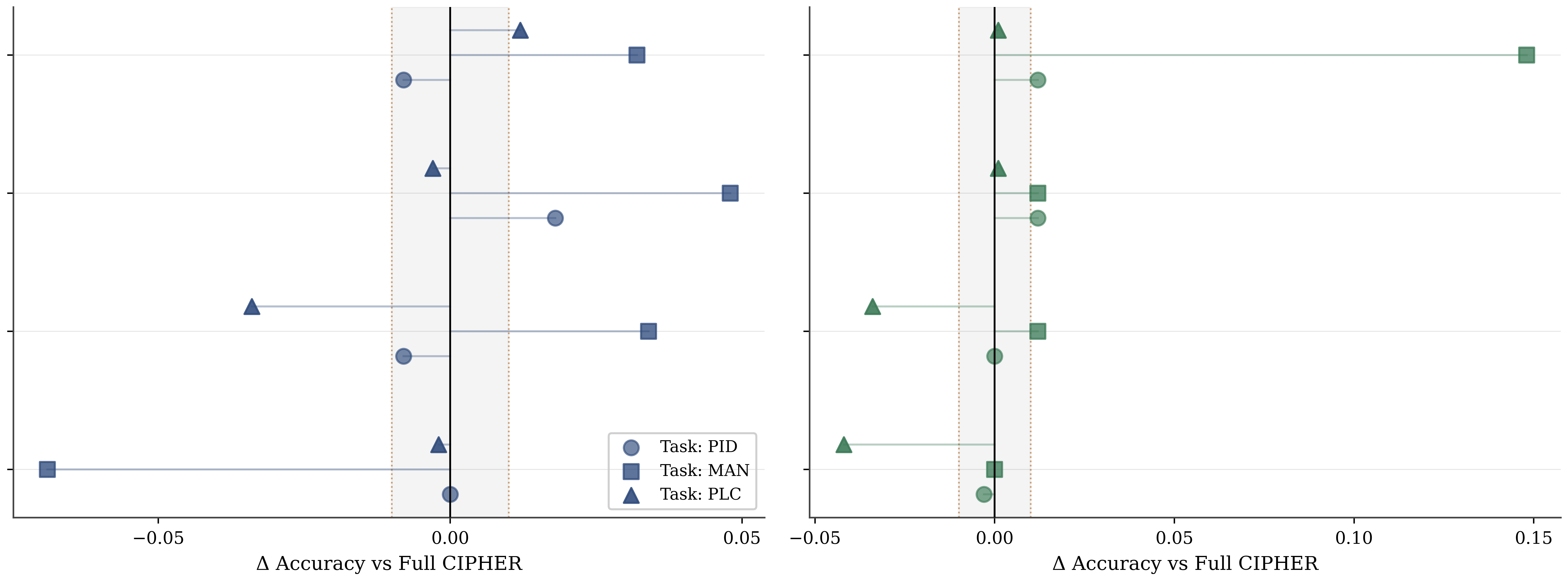}
}
\caption{Ablation deltas relative to full CIPHER across feature-task settings.}
\label{fig:ablation_deltas}
\end{figure}

To close the previous replication gap, we ran the expanded ablation matrix with
three seeds (17, 31, 53) across ERP and DDA for phoneme identity, manner, and
place on the 8-8 split (Figures~\ref{fig:ablation_heatmap}
and~\ref{fig:ablation_deltas}).

\begin{table}[H]
\centering
\caption{Multi-seed ablation matrix (8-8 split, 3 seeds).}
\label{tab:ablation_full}
\small
\begin{tabular}{llccc}
\toprule
\textbf{Feature} & \textbf{Task} & \textbf{Best Variant (acc)} &
  \textbf{Full CIPHER} & \textbf{Largest $-\Delta$ vs Full} \\
\midrule
ERP & Phoneme identity & No attn pool (0.150)  & 0.132 & No multi-scale ($-0.008$) \\
ERP & Manner           & No attn pool (0.822)  & 0.774 & No SE ($-0.069$)          \\
ERP & Place            & No multi-scale (0.563)& 0.551 & No stoch.\ depth ($-0.034$)\\
DDA & Phoneme identity & No multi-scale (0.137)& 0.125 & No SE ($-0.003$)          \\
DDA & Manner           & No multi-scale (0.852)& 0.704 & None observed             \\
DDA & Place            & No attn pool (0.572)  & 0.571 & No SE ($-0.042$)          \\
\bottomrule
\end{tabular}
\end{table}

\textbf{Key Findings:}
\begin{itemize}
  \item Multi-seed replication confirms that component effects are strongly
    task-dependent rather than globally monotonic.
  \item SE removal is the most consistently harmful perturbation across several
    settings (notably ERP manner and DDA place), while stochastic depth has
    mixed effects by task.
  \item The replicated matrix supports an ``engineering trade-off'' claim more
    than a single dominant architectural mechanism claim.
\end{itemize}

\section{Discussion}

\subsection{Perfect Articulatory Classification: Signal or Artifact?}

The 100\% accuracy on binary articulatory features warrants strict skepticism.
Several factors can fully account for this result without requiring successful
decoding of abstract speech representations:

\begin{enumerate}
  \item \textbf{Acoustic-ERP confounds.} Manner and place distinctions
    correspond to dramatically different acoustic onsets (stop burst
    vs.\ fricative noise), which produce distinct auditory ERPs (P1/N1/P2
    complex) even before motor-speech processing occurs. The model may be
    exploiting these early auditory-cortical responses rather than higher-order
    phonological representations; this concern is consistent with prior
    speech-ERP analyses of acoustic-feature
    separability~\cite{toscano2010}.
  \item \textbf{Binary classification ceiling.} With only 2 classes and a
    strong auditory ERP, a well-regularized deep model can easily achieve
    near-perfect performance. The more informative metric is the 11-class
    phoneme identity task, where accuracy is substantially lower.
  \item \textbf{TMS-target confounding by design.} In this paradigm, TMS
    target and phoneme category are partially blocked together. Even after
    artifact cleaning, residual pulse-related transients or downstream
    perturbation signatures can remain correlated with class labels, yielding
    inflated decoding.
\end{enumerate}

Given these issues, binary-task ceiling results should not be used as primary
evidence for neural speech decoding. They are better treated as
confound-sensitive upper bounds pending dedicated control analyses.

\subsection{ERP-DDA Trade-offs for Fine-Grained Decoding}

Under full Study~2 LOSO, ERP and DDA show mixed performance on triphone WER
rather than a single dominant winner (real words: ERP 0.671 vs.\ DDA 0.688;
pseudowords: ERP 0.780 vs.\ DDA 0.772). This pattern should be interpreted as
a feature trade-off in a difficult constrained classification setup rather than
evidence of robust EEG-to-text decoding. DDA coefficients operate on the raw
2048\,Hz signal and may capture broader dynamical structure, including:

\begin{itemize}
  \item High-frequency gamma oscillations (30--100+\,Hz) linked to speech
    processing
  \item Cross-frequency coupling patterns that index phonological processing
  \item Attractor geometry differences between phoneme categories
\end{itemize}

In contrast, the ERP pathway applies a 40\,Hz low-pass filter, discarding
higher-frequency content that may carry additional discriminative cues. In our
pooled-feature LR control, a wider-band ERP variant (0.5--100\,Hz) did not
improve LOSO summary metrics over 0.5--40\,Hz, but this does not rule out
potential differences in end-to-end deep models.

\subsection{Cross-Dataset Generalization}

Models trained on Study~2 transfer to Study~1 with high binary-task
performance. However, because the dominant cues in these tasks may be acoustic
and/or TMS-related, transfer alone cannot be taken as evidence of universal
neural speech representations.

The cross-study transfer remains useful as a robustness check, but its
scientific interpretation depends on ruling out nuisance cues with explicit
negative controls.

\subsection{Limitations and Remaining Gaps}

\begin{enumerate}
  \item \textbf{Sample size.} With only 24 participants across both studies,
    statistical power for individual-level analyses (e.g., TMS effects,
    lexicality) is limited.
  \item \textbf{Perception only.} The dataset contains auditory perception
    trials, not overt or imagined production; this limits clinical BCI
    relevance.
  \item \textbf{Task scope.} The system performs stimulus-locked, labeled
    classification, not open-vocabulary text generation.
  \item \textbf{Confound exposure remains material.} The completed NULL-only
    and acoustic controls confirm that nuisance cue pathways can dominate
    decoding outcomes, especially on binary tasks.
  \item \textbf{Baseline scope remains limited but expanded.} We now include
    matched-split baselines (chance, LR, LDA, ShallowConvNet, EEGNet,
    EEG-Conformer), but additional comparators and stricter
    architecture-matched tuning remain future work.
  \item \textbf{Ablation evidence is now multi-seed and broader.} We report
    three-seed ablations across ERP and DDA for phoneme identity, manner, and
    place; remaining uncertainty is due to dataset scale and optimization
    variance, not missing replication coverage.
  \item \textbf{Offline analysis.} All results are offline; real-time
    viability remains untested.
\end{enumerate}

\subsection{Confound Isolation: What the Control Suite Establishes}

The central interpretive risk in stimulus-locked EEG decoding is that
apparent classification gains reflect acoustic or experimental-design
artefacts rather than neural speech representations. To bound this risk,
we executed a pre-specified set of controls whose results, taken together,
sharpen the scope of admissible claims.

Training and evaluating exclusively on NULL-condition trials under
leave-one-subject-out cross-validation removes TMS-linked label
correlations entirely. Under this stricter protocol, pooled-feature EEG
baselines reach 0.104 (ERP) and 0.129 (DDA) on 11-class phoneme identity
--- well above chance (0.091), but far below the ceiling figures reported
on mixed-condition splits. The gap directly quantifies how much of the
earlier apparent performance was absorbed by TMS-target blocking rather
than phoneme-discriminative EEG structure.

The acoustic control is the sharpest result in the suite: metadata-derived
stimulus features alone achieve perfect LOSO accuracy on every binary
articulatory task. This establishes that manner, place, voicing, and
category labels are fully recoverable from the stimulus identity without
any EEG signal at all. Consequently, binary-task EEG results carry no
evidential weight for neural decoding --- they are reported as
upper-bound sanity checks, not as scientific claims.

Masking the early auditory window (0--200\,ms) produces only small
accuracy shifts (manner $-0.025$, place $-0.005$, phoneme $-0.0003$),
indicating that the pooled-feature control models draw on the full epoch
rather than exclusively on the initial transient. Block-aware label
permutation within TMS conditions yields non-significant empirical
$p$-values across all feature--task combinations ($p \geq 0.529$),
confirming that the observed LOSO accuracies are not an artefact of
within-block label structure.

Taken together, these controls do not establish that CIPHER decodes
phonological representations --- they establish the \emph{conditions
under which that claim would need to be evaluated}. The one result that
survives this scrutiny is the 11-class CVC phoneme WER under NULL-only
LOSO, where acoustic and TMS confounds are jointly suppressed. That
metric, and not the binary articulatory accuracies, is the appropriate
basis for any further scientific interpretation.

\section{Architectural Details and Hyperparameters}

Table~\ref{tab:hyperparams} provides the full hyperparameter specification for
reproducibility.

\begin{table}[H]
\centering
\caption{Full hyperparameter specification.}
\label{tab:hyperparams}
\begin{tabular}{ll}
\toprule
\textbf{Hyperparameter} & \textbf{Value} \\
\midrule
Model dimension $d_{\text{model}}$ & 256 \\
Conformer blocks                   & 4 \\
Attention heads                    & 8 \\
Conv front-end channels            & 64 per branch (192 total) \\
Conv front-end kernels             & $\{3, 7, 15\}$ \\
Conformer conv kernel              & 15 \\
SE reduction ratio                 & 4 \\
Dropout                            & 0.2 \\
Drop path rate                     & 0.05 (linearly increased across blocks) \\
Label smoothing $\epsilon$         & 0.1 \\
Mixup $\alpha$                     & 0.1 (disabled for last 10\% of epochs) \\
Learning rate                      & $5 \times 10^{-4}$ \\
Weight decay                       & $10^{-4}$ \\
Optimizer                          & AdamW ($\beta_1 = 0.9$, $\beta_2 = 0.98$) \\
Batch size                         & 64 \\
Max epochs                         & 150 \\
Early stopping patience            & 30 epochs \\
Warmup epochs                      & 10 \\
Gradient clipping                  & Max-norm 1.0 \\
CTC weight                         & 0.1 \\
ERP sampling rate                  & 256\,Hz \\
DDA sampling rate                  & 2048\,Hz \\
DDA window / shift                 & 60 / 2 samples \\
DDA delays $(\tau_1, \tau_2)$      & 6, 16 samples \\
DDA temporal stride                & 4 \\
Epoch window                       & $[-200, +800]$\,ms \\
Augmentation: noise $\sigma$       & $0.02 \times$ sample std \\
Augmentation: channel dropout      & 5--10\% \\
Augmentation: time shift           & $\pm$5 (ERP) / $\pm$20 (DDA) samples \\
Augmentation: time mask            & 5--10\% of time steps \\
Augmentation: amplitude scale      & $[0.85, 1.15]$ \\
\bottomrule
\end{tabular}
\end{table}

\section{Conclusion}

CIPHER establishes a dual-pathway benchmark for stimulus-locked EEG phoneme
decoding, pairing a Conformer-based encoder with two complementary feature
representations---narrowband ERPs and broadband DDA coefficients---and
subjecting both to a complete pre-specified confound-control suite. The
central finding is that binary articulatory classification, while
near-ceiling, is entirely explained by acoustic onset separability and
TMS-target blocking, and carries no independent evidential weight for neural
speech-feature decoding. The one result that survives all controls
simultaneously is the 11-class CVC phoneme WER under NULL-only LOSO (best:
ERP $0.671 \pm 0.080$ on real words), which is above chance but leaves
substantial error, and should be interpreted as a measure of EEG
discriminability under constrained conditions rather than as a practical
decoding capability.

ERP and DDA show complementary lexicality profiles---ERP stronger on real
words, DDA stronger on pseudowords---with neither pathway dominating
consistently. Ablations identify SE channel attention as the most reliable
architectural contributor, while multi-scale branching and stochastic depth
show task-dependent effects that fall within optimization variance at current
dataset scale.

The primary value of this work is methodological: a replicable benchmark
protocol, a transparent treatment of confound exposure, and a
feature-comparison framework applicable to future datasets with larger
participant pools, imagined speech paradigms, and extended vocabulary.
Closing the gap between the WER reported here and practically useful
decoding will require all three.

\section*{Declaration of Generative AI and AI-Assisted Technologies}
\addcontentsline{toc}{section}{Declaration of Generative AI and AI-Assisted Technologies}

\begin{mdframed}[backgroundcolor=gray!8, linecolor=gray!40, linewidth=0.7pt,
                 innerleftmargin=12pt, innerrightmargin=12pt,
                 innertopmargin=10pt, innerbottommargin=10pt]
The following AI tools were used during the preparation of this work:

\begin{itemize}[leftmargin=1.4em, itemsep=3pt]
  \item \textbf{Claude} (Anthropic) --- used for assistance in documenting
    code and generating plotting code for selected figures.
  \item \textbf{ChatGPT} (OpenAI) --- used for assistance in documenting
    code and generating plotting code for selected figures.
  \item \textbf{GitHub Copilot} (OpenAI GPT-5.3 Codex / Claude Opus~4.6) ---
    used for generating plotting code for selected figures and for code
    documentation assistance.
\end{itemize}

All scientific content, experimental design, analysis, interpretation of
results, and conclusions are the sole responsibility of the author. AI-generated
content was reviewed, edited, and verified before inclusion. No AI tool was used
to generate, fabricate, or alter experimental data or results.
\end{mdframed}

\bibliographystyle{unsrt}
\bibliography{references}

@article{willett2023,
  author  = {Willett, Frank R. and Kunz, Erin M. and Fan, Chaofei and
             Avansino, Donald T. and Wilson, Guy H. and Choi, Eun Young and
             Kamdar, Foram and Hochberg, Leigh R. and Druckmann, Shaul and
             Shenoy, Krishna V. and Henderson, Jaimie M.},
  title   = {A high-performance speech neuroprosthesis},
  journal = {Nature},
  year    = {2023},
  volume  = {620},
  pages   = {1031--1036},
  doi     = {10.1038/s41586-023-06377-x}
}

@article{metzger2023,
  author  = {Metzger, Sean L. and Liu, Jessie R. and Moses, David A. and
             Dougherty, Matthew E. and Liu, Margaret P. and Bhaya-Grossman, Ilina and
             Burkhart, Michelle C. and Bhaskaran, Maitreyee and Friedenberg, David A. and
             Osborn, Laura E. and Ganguly, Karunesh and Chang, Edward F.},
  title   = {A high-performance neuroprosthesis for speech decoding and avatar control},
  journal = {Nature},
  year    = {2023},
  volume  = {620},
  pages   = {1037--1046},
  doi     = {10.1038/s41586-023-06443-4}
}

@article{defossez2023,
  author  = {Défossez, Alexandre and Caucheteux, Charlotte and Rapin, Jérémy and
             Kabeli, Ori and King, Jean-Rémi},
  title   = {Decoding speech perception from non-invasive brain recordings},
  journal = {Nature Machine Intelligence},
  year    = {2023},
  volume  = {5},
  pages   = {1097--1107},
  doi     = {10.1038/s42256-023-00714-5}
}

@article{anumanchipalli2019,
  author  = {Anumanchipalli, Gopala K. and Chartier, Josh and Chang, Edward F.},
  title   = {Speech synthesis from neural decoding of spoken sentences},
  journal = {Nature},
  year    = {2019},
  volume  = {568},
  pages   = {493--498},
  doi     = {10.1038/s41586-019-1119-1}
}

@article{lainscsek2015,
  author  = {Lainscsek, Claudia and Sejnowski, Terrence J.},
  title   = {Delay differential analysis of time series},
  journal = {Neural Computation},
  year    = {2015},
  volume  = {27},
  number  = {3},
  pages   = {594--614},
  doi     = {10.1162/NECO_a_00706}
}

@inproceedings{gulati2020,
  author    = {Gulati, Anmol and Qin, James and Chiu, Chung-Cheng and
               Parmar, Niki and Zhang, Yu and Yu, Jiahui and Han, Wei and
               Wang, Shibo and Zhang, Zhengdong and Wu, Yonghui and Pang, Ruoming},
  title     = {Conformer: Convolution-augmented {T}ransformer for speech recognition},
  booktitle = {Proceedings of Interspeech},
  year      = {2020},
  pages     = {5036--5040},
  doi       = {10.21437/Interspeech.2020-3015}
}

@article{herff2016,
  author  = {Herff, Christian and Schultz, Tanja},
  title   = {Automatic speech recognition from neural signals: {A} focused review},
  journal = {Frontiers in Neuroscience},
  year    = {2016},
  volume  = {10},
  pages   = {429},
  doi     = {10.3389/fnins.2016.00429}
}

@article{schirrmeister2017,
  author  = {Schirrmeister, Robin Tibor and Springenberg, Jost Tobias and
             Fiederer, Lukas Dominique Josef and Glasstetter, Martin and
             Eggensperger, Katharina and Tangermann, Michael and Hutter, Frank and
             Burgard, Wolfram and Ball, Tonio},
  title   = {Deep learning with convolutional neural networks for {EEG} decoding
             and visualization},
  journal = {Human Brain Mapping},
  year    = {2017},
  volume  = {38},
  number  = {11},
  pages   = {5391--5420},
  doi     = {10.1002/hbm.23730}
}

@article{song2023,
  author  = {Song, Yonghao and Zheng, Qingqing and Liu, Bingchuan and Gao, Xiaorong},
  title   = {{EEG} {C}onformer: Convolutional {T}ransformer for {EEG} decoding
             and visualization},
  journal = {IEEE Transactions on Neural Systems and Rehabilitation Engineering},
  year    = {2023},
  volume  = {31},
  pages   = {710--719},
  doi     = {10.1109/TNSRE.2022.3230250}
}

@article{lawhern2018,
  author  = {Lawhern, Vernon J. and Solon, Amelia J. and Waytowich, Nicholas R. and
             Gordon, Stephen M. and Hung, Chou P. and Lance, Brent J.},
  title   = {{EEGNet}: A compact convolutional neural network for {EEG}-based
             brain--computer interfaces},
  journal = {Journal of Neural Engineering},
  year    = {2018},
  volume  = {15},
  number  = {5},
  pages   = {056013},
  doi     = {10.1088/1741-2552/aace8c}
}

@inproceedings{duan2023,
  author    = {Duan, Yiqun and Zhou, Charles and Wang, Zhen and Wang, Yu-Kai and
               Lin, Chin-Teng},
  title     = {{DeWave}: Discrete encoding of {EEG} waves for {EEG} to text
               translation},
  booktitle = {Advances in Neural Information Processing Systems (NeurIPS)},
  year      = {2023},
  volume    = {36}
}

@article{lainscsek2017,
  author  = {Lainscsek, Claudia and Hernandez, Manuel Enrique and Weyhenmeyer,
             Jonathan and Sejnowski, Terrence J. and Poizner, Howard},
  title   = {Delay differential analysis of seizures in multichannel
             electrocorticography data},
  journal = {Neural Computation},
  year    = {2017},
  volume  = {29},
  number  = {12},
  pages   = {3181--3218},
  doi     = {10.1162/neco_a_01009}
}

@article{liberman1985,
  author  = {Liberman, Alvin M. and Mattingly, Ignatius G.},
  title   = {The motor theory of speech perception revised},
  journal = {Cognition},
  year    = {1985},
  volume  = {21},
  number  = {1},
  pages   = {1--36},
  doi     = {10.1016/0010-0277(85)90021-6}
}

@article{dausilio2009,
  author  = {D'Ausilio, Alessandro and Pulvermüller, Friedemann and Salmas, Paola and
             Bufalari, Ilaria and Begliomini, Chiara and Fadiga, Luciano},
  title   = {The motor somatotopy of speech perception},
  journal = {Current Biology},
  year    = {2009},
  volume  = {19},
  number  = {5},
  pages   = {381--385},
  doi     = {10.1016/j.cub.2009.01.017}
}

@misc{moreira2024,
  author       = {Moreira, Jo\~{a}o P. C. and others},
  title        = {An open-access {EEG} dataset for speech decoding: {E}xploring
                  the role of articulation and coarticulation},
  howpublished = {OpenNeuro ds006104},
  year         = {2025},
  url          = {https://openneuro.org/datasets/ds006104}
}

@inproceedings{hu2018,
  author    = {Hu, Jie and Shen, Li and Sun, Gang},
  title     = {Squeeze-and-excitation networks},
  booktitle = {Proceedings of the IEEE Conference on Computer Vision and
               Pattern Recognition (CVPR)},
  year      = {2018},
  pages     = {7132--7141},
  doi       = {10.1109/CVPR.2018.00745}
}

@inproceedings{graves2006,
  author    = {Graves, Alex and Fern\'{a}ndez, Santiago and Gomez, Faustino and
               Schmidhuber, J\"{u}rgen},
  title     = {Connectionist temporal classification: {L}abelling unsegmented
               sequence data with recurrent neural networks},
  booktitle = {Proceedings of the International Conference on Machine Learning
               (ICML)},
  year      = {2006},
  pages     = {369--376},
  doi       = {10.1145/1143844.1143891}
}

@inproceedings{zhang2018,
  author    = {Zhang, Hongyi and Ciss\'{e}, Moustapha and Dauphin, Yann N. and
               Lopez-Paz, David},
  title     = {Mixup: {B}eyond empirical risk minimization},
  booktitle = {Proceedings of the International Conference on Learning
               Representations (ICLR)},
  year      = {2018}
}

@article{toscano2010,
  author  = {Toscano, Joseph C. and McMurray, Bob},
  title   = {Cue integration with categories: {W}eighting acoustic cues in
             speech using unsupervised learning and distributional statistics},
  journal = {Cognitive Science},
  year    = {2010},
  volume  = {34},
  number  = {3},
  pages   = {434--464},
  doi     = {10.1111/j.1551-6709.2009.01077.x}
}

@article{combrisson2015,
  author  = {Combrisson, Etienne and Jerbi, Karim},
  title   = {Exceeding chance level by chance: {T}he caveat of theoretical
             chance levels in brain signal classification and statistical
             assessment of decoding accuracy},
  journal = {Journal of Neuroscience Methods},
  year    = {2015},
  volume  = {250},
  pages   = {126--136},
  doi     = {10.1016/j.jneumeth.2015.01.010}
}

@article{doi:10.1056/NEJMoa2314132,
author = {Nicholas S. Card  and Maitreyee Wairagkar  and Carrina Iacobacci  and Xianda Hou  and Tyler Singer-Clark  and Francis R. Willett  and Erin M. Kunz  and Chaofei Fan  and Maryam Vahdati Nia  and Darrel R. Deo  and Aparna Srinivasan  and Eun Young Choi  and Matthew F. Glasser  and Leigh R. Hochberg  and Jaimie M. Henderson  and Kiarash Shahlaie  and Sergey D. Stavisky  and David M. Brandman },
title = {An Accurate and Rapidly Calibrating Speech Neuroprosthesis},
journal = {New England Journal of Medicine},
volume = {391},
number = {7},
pages = {609-618},
year = {2024},
doi = {10.1056/NEJMoa2314132},

URL = {https://www.nejm.org/doi/full/10.1056/NEJMoa2314132},
eprint = {https://www.nejm.org/doi/pdf/10.1056/NEJMoa2314132}
}

\appendix

\end{document}